\pdfoutput=1

\documentclass[11pt]{article}

\usepackage{naacl2021}

\usepackage{times}
\usepackage{latexsym}
\usepackage{graphicx}
\usepackage{algorithmic}
\usepackage{amsfonts}
\usepackage[scientific-notation=true]{siunitx}
\usepackage[bottom]{footmisc}

\usepackage[T1]{fontenc}

\usepackage[utf8]{inputenc}

\usepackage{microtype}

\usepackage[ruled,vlined]{algorithm2e}

\title{On Geodesic Distances and Contextual Embedding Compression for Text Classification}

\author{Rishi Jha\thanks{\text{ } Equal contribution}\text{ }  \and Kai Mihata\footnotemark[1]\\
        Paul G. Allen School of Computer Science \& Engineering, \\ University of Washington, Seattle, WA, USA \\
        \texttt{\{rjha01, kaim2\}@cs.washington.edu}}

\begin{document}
\maketitle
\begin{abstract}
In some memory-constrained settings like IoT devices and over-the-network data pipelines, it can be advantageous to have smaller contextual embeddings. We investigate the efficacy of projecting contextual embedding data (BERT) onto a manifold, and using nonlinear dimensionality reduction techniques to compress these embeddings. In particular, we propose a novel post-processing approach, applying a combination of Isomap and PCA. We find that the geodesic distance estimations, estimates of the shortest path on a Riemannian manifold, from Isomap's k-Nearest Neighbors graph bolstered the performance of the compressed embeddings to be comparable to the original BERT embeddings. On one dataset, we find that despite a 12-fold dimensionality reduction, the compressed embeddings performed within 0.1\% of the original BERT embeddings on a downstream classification task. In addition, we find that this approach works particularly well on tasks reliant on syntactic data, when compared with linear dimensionality reduction. These results show promise for a novel geometric approach to achieve lower dimensional text embeddings from existing transformers and pave the way for data-specific and application-specific embedding compressions. 

\end{abstract}

\begin{figure*}[t]
\includegraphics[width=8cm]{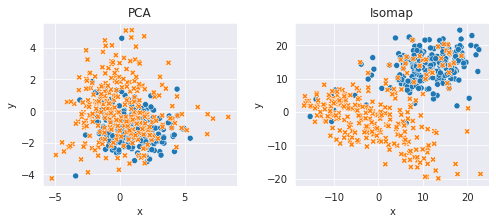}
\centering
\caption{Visualization of two-dimensional PCA and Isomap compressions based on BERT embeddings for the SMS-SPAM dataset \citep{Almeida2013TowardsSS}. Spam is represented by a blue dot and ham by an orange x. We see that for this dataset in two dimensions, the Isomap compression appears more linearly separable than the PCA compression, making classification easier for the former.}
\label{fig:spam}
\end{figure*}

\section{Introduction}

Contextual embeddings, like those BERT \citep{devlin-etal-2019-bert} generates, improve on non-contextual word embeddings by providing contextual semantics to the real-valued representation of a text. Although these models have been shown to achieve state-of-the-art performance on most NLP tasks, they are notably expensive to train. To help combat this, as mentioned by \citet{2019arXiv190901264M}, model compression techniques like data quantization \citep{2014arXiv1412.6115G}, model pruning \citep{han2015deep_compression}, and knowledge distillation (\citealp{Sanh2019DistilBERTAD}, \citealp{44873}) have been developed. However, at 768 dimensions, the embeddings themselves can be prohibitively large for some tasks and settings.

Smaller embeddings both enable more compact data sizes in storage-constrained settings and over-the-air data pipelines, and help lower the requisite memory for using the embeddings for downstream tasks. For non-contextual word embeddings, \citet{ling-etal-2016-word} note that loading matrices can take multiple gigabytes of memory, a prohibitively large amount for some phones and IoT devices. While contextual embeddings are smaller, downstream models will face similar headwind for large corpora.

Although there has been more extensive study in the efficacy of compressing non-contextual word embeddings (\citealp{raunak-etal-2019-effective}, \citealp{784d5b21fdb248c588a0ea2d4443e6af}), to the best of our knowledge few contextual embedding compression post-processing approaches have been proposed \citep{li-eisner-2019-specializing}. In their work, \citet{li-eisner-2019-specializing} propose the Variational Information Bottleneck, an autoencoder to create smaller, task specific embeddings for different languages. While effective, the computational expense of additional training loops is not appropriate for some memory constrained applications.

Our approach more closely mirrors the work of \citet{raunak-etal-2019-effective} who propose a Principal Component Analysis (PCA)-based post-processing algorithm to lower the dimensionality of non-contextual word embeddings. They find that they can replicate, or, in some cases, increase the performance of the original embeddings. One limitation to this approach is the lack of support for nonlinear data patterns. Nonlinear dimensionality reductions, like the Isomap shown in Figure~\ref{fig:spam}, can pick up on latent textual features that evade linear algorithms like PCA. To achieve this nonlinearity, we extend this approach to contextual embeddings, adding in additional geodesic distance information via the Isomap algorithm \citep{Tenenbaum2319}. To the best of our knowledge, the application of graph-based techniques to reduce the dimensionality of contextual embeddings is novel.

The goal of this paper is not to compete with state-of-the-art models, but, rather, (1) to show that 12-fold dimensionality reductions of contextual embeddings can, in some settings, conserve much of the original performance, (2) to illustrate the efficacy of geodesic similarity metrics in improving the downstream performance of contextual embedding compressions, and (3) propose the creation of more efficient, geodesic-distance-based transformer architectures. In particular, our main result is showing that a 64-dimensional concatenation of compressed PCA and Isomap embeddings are comparable to the original BERT embeddings and outperform our PCA baseline. We attribute this success to the locality data preserved by the k-Nearest Neighbors (k-NN) graph generated by the Isomap algorithm.

\begin{figure*}[t]
\includegraphics[width=8cm]{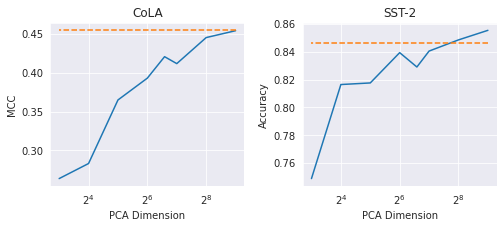}
\centering
\caption{PCA baseline performance on CoLA \citep{warstadt-etal-2019-neural} and SST-2 \citep{socher-etal-2013-recursive}. PCA embedding performance by dimension is represented by the solid blue line. Regression at 768 dimensions is represented by an orange dashed line. On these two datasets, even at much smaller dimensionality, we see that PCA has comparable performance. }
\label{fig:performance}
\end{figure*}

\section{Related Work}

As best we know, there is very little literature regarding the intersection of contextual embedding compression and geodesic distances. Most of the existing work in related spaces deals with non-contextual word embeddings. Despite the rapid growth in the popularity of transformers, these embeddings still retain popularity.

For non-contextual word embeddings, \citet{784d5b21fdb248c588a0ea2d4443e6af} propose a post-processing algorithm that projects embedded data away from the dominant principal components, in order to greater differentiate the data. \citet{raunak-etal-2019-effective} expand on this algorithm by combining it with PCA reductions. Both approaches are effective, but, are limited to linear dimensionality reductions. 

Some nonlinear approaches include \citet{10.1007/978-3-319-46681-1_50} and \citet{li-eisner-2019-specializing} who both use autoencoder-based compressions. Notably, the former only addresses non-contextual embeddings.

Meanwhile the usage of graphs in NLP is well established, but their usage in the compression of contextual embeddings is not well documented. \citet{wiedemann-etal-2019-does} use a k-NN classification to achieve state-of-the-art word sense disambiguation. Their work makes clear the effectiveness of the k-NN approach in finding distinctions in hyper-localized data.

\section{Method}
\label{sec:method}
With the goal of reducing the contextual embedding dimensionality, we first processed our data using a pre-trained, uncased BERT Base model. Then, we compressed the data to a lower dimension using both PCA and Isomap as described in Section~\ref{sec:experiments}. This method aims to capture as much information as possible from the original BERT embeddings while preserving graphical locality information and nonlinearities in the final contextual embeddings.

\subsection{Isomap and Geodesic Distances}
For this paper, to blend geodesic distance information and dimensionality reduction, we use Tenenbaum's Isomap \citep{Tenenbaum2319}. Isomap relies on a weighted neighborhood graph that allows for the inclusion of complex, nonlinear patterns in the data, unlike a linear algorithm like PCA. In specific, this graph is constructed so that the edges between each vertex (datapoint) and its k-nearest neighbors have weight corresponding to the pairwise Euclidean distance on a Riemannian manifold. Dijkstra's shortest path between two points then estimates their true geodesic distance. These geodesics are particularly useful for delineating points that are close in Euclidean space but not on a manifold i.e. similar BERT embeddings with different meanings.

If we assume the data follows a manifold, Isomap can exploit the Riemannian locality of these complex contextual embeddings. As Figure~\ref{fig:spam} shows, in some cases this is a good assumption to make since we are then able to dissect complex embeddings into near-linearly separable clusters. Notably, there are some limitations to this approach. If the manifold is sparse, i.e. there are few data points on certain regions of the manifold, or k is too small, the shortest path estimation of the geodesic distance can be unrepresentative of the true distance. On the contrary, if k is too large, Isomap overgeneralizes and loses its fine-grained estimation of the Riemannian surface.

Nonetheless, we hypothesize that these global geodesic distance approximations explain the empirical advantage Isomap has in our setting over other popular nonlinear dimensionality reduction techniques. Many alternatives, like Locally Linear Embeddings \citep{Roweis2000} focus, instead, on preserving intra-neighborhood distance information that may not encompass inter-neighborhood relationships as Isomap does.

\subsection{Our Approach}
We applied our post-processing method to the BERT embeddings through three different dimensionality reductions. We used (1) PCA, (2) Isomap, and (3) a concatenation of embeddings from the two before training a small regression model on the embeddings. This approach aims to use linear and nonlinear dimensionality reduction techniques to best capture the data's geodesic locality information.
\label{sec:experiments}

\paragraph{PCA.}
To compute linearly-reduced dimensionality embeddings, we used PCA to reduce the 768-dimensional BERT embeddings down to a number of components ranging from 16 to 256. While there are other linear dimensionality reduction techniques, PCA is a standard benchmark and empirically performed the best. These serve as a linear baseline for reduced dimension embeddings.

\paragraph{Isomap.}
To compute geodesic locality information, we post-processed our BERT embeddings with Isomap. The final Isomap embeddings ranged from 16 to 96 dimensions, all computed with 96 neighbors and Euclidean distance. 

\paragraph{Concatenated Embeddings.}
To include features from both of these reductions, we combined an Isomap embedding with a PCA embedding to form concatenations of several dimensions. We experimented with ratios of PCA embedding size to Isomap embedding size from 0 to \(\frac12\) at \(\frac18\) intervals. We found that this ratio was the main determinant of relative accuracy, so for analysis we fixed the total dimension to 64.
\begin{table*}
\centering
\begin{tabular}{lccccc}
\hline
\textbf{Embedding} & \textbf{Isomap Dim} & \textbf{PCA Dim} & \textbf{CoLA} & \textbf{SST-2}\\
\hline
PCA & N/A & 64 & 0.339 & 0.842  \\
Concatenation* & 16 & 48 & 0.421 & 0.846  \\
Concatenation & 32 & 32 & 0.384 & 0.822  \\
Concatenation & 48 & 16 & 0.357 & 0.817  \\
Isomap & 64 & N/A & 0.332 & 0.814  \\
BERT & N/A & N/A & 0.455 & 0.847\\
\hline
\end{tabular}
\caption{64-dimensional embedding performance on CoLA \citep{warstadt-etal-2019-neural} and SST-2 \citep{socher-etal-2013-recursive}. CoLA is measured by Matthews correlation and SST-2 by accuracy. While Isomap did not perform the best outright, on these datasets we found that some inclusion of locality data proved meaningful. This shows the trade-off between locality information and performance mentioned in Section~\ref{sec:concatembed}. The best 12-fold compression performance is asterisked.}
\label{tab:concat}
\end{table*}

\section{Experiments and Results}
\label{sec:exp}
We assess the results of these compression techniques on two text classification datasets. We provide the code for our experiments\footnote{\href{https://github.com/kaimihata/geo-bert}{https://github.com/kaimihata/geo-bert}}.

\subsection{Data}
We evaluate our method on two text classification tasks: the Corpus of Linguistic Acceptability (CoLA) \citep{warstadt-etal-2019-neural}, a 10657 sentence binary classification dataset on grammatical correctness and the Stanford Sentiment Treebank v2 (SST-2) \citep{socher-etal-2013-recursive}, a 70042 sentence binary (positive / negative) sentiment classification dataset.

For CoLA, we used the predefined, 9594 datapoint train set and for SST-2, we used the first 8000 samples of their training set to construct ours due to computational limitations. For testing and evaluation, we used the corresponding datasets defined by GLUE \citep{wang-etal-2018-glue}. In addition, for all of our evaluations, we used the same pre-trained BERT embeddings for consistency.

\subsection{Training and Evaluation}
All of these post-processed embeddings, as well as the BERT embeddings, were trained on a downstream regression model consisting of one hidden layer (64 dim) with ReLU activation, a learning rate of \num{0.0001}, and were optimized via ADAM \citep{DBLP:journals/corr/KingmaB14}. The BERT embeddings are used as a baseline for comparison. 

To evaluate our embeddings on CoLA and SST-2, we used their GLUE-defined metrics of Matthews correlation and validation accuracy, respectively. For each embedding experiment, our procedure consisted of running our post-processing method on the BERT embeddings then training the downstream model. Each reported metric is the average of three of these procedures.

\subsection{Baseline Comparison}
Agnostic of post-processing algorithm, we found reduced-dimensionality embeddings were competitive with the original embeddings. Although smaller reduction factors, understandably, performed better, we found that even when reduced by a factor as large as 12, our PCA embeddings experienced small losses in performance on both datasets (Figure~\ref{fig:performance}). To demonstrate the effect of the inclusion of locality data, we picked an embedding size of 64 dimensions (a reduction factor of 12) to balance embedding size and performance for our main experiment.

In comparison to our 768-dimensional baseline, at 64 dimensions, the best reduction results were within 7.5\% and 0.1\% for CoLA and SST-2, respectively (Table~\ref{tab:concat}). These results show that with or without the presence of locality data, compressed embeddings can perform comparably to the original embeddings. 

\subsection{Locality Information Trade-off}
\label{sec:concatembed}

As shown in Table~\ref{tab:concat}, on neither dataset did the fully PCA or Isomap embeddings perform the best. The best performer was, instead, a combination of these two approaches. This indicates that there must exist a trade-off on the effectiveness of locality data. While without locality data, the embedding obviously misses out on geodesic relationships, too much locality information may replace more useful features that the PCA embeddings extract. Just as the quality of the geodesic distance estimations rely on how well the data fits the underlying manifold, as discussed in Section~\ref{sec:method}, so, too, does its effectiveness. To explain this phenomenon, we hypothesize that the addition of small amounts of locality data bolsters performance by describing the geodesic relationships without drowning out important syntactic and semantic information provided by PCA.

\subsection{Task-Specific Locality}

While the best reduction consisted of a concatenation of 16-dimensional Isomap and 48-dimensional PCA embeddings, whether the other concatenations performed better than our PCA baseline was dependent on the task. For CoLA, we found that all three concatenated embeddings performed better than PCA, whereas for SST-2, only the top performing concatenated embedding beat out our baseline. To describe this disparity we look towards the nature of the datasets and tasks. Notably, CoLA requires models to identify proper grammar, a syntactic task, while SST-2 requires models to understand the sentiment of sentences, a semantic task. Syntactic data often has some intrinsic structure to it, and perhaps our manifold approach encompasses this information well. Based on this result, exploring this distinction could be an exciting avenue for further study.

\section{Conclusions and Future Work}

We present a novel approach for compressing BERT embeddings into effective lower dimension representations. Our method shows promise for the inclusion of geodesic locality information in transformers and future compression methods. We hope our results lead to more work investigating the geometric structure of transformer embeddings and developing more computationally efficient NLP training pipelines. To further this work, we plan to investigate the efficacy of (1) other graph dimensionality reduction techniques, (2) non-Euclidean distance metrics, and (3) our approach on different transformers. In addition, we would like to investigate whether datasets for other tasks can be effectively projected onto a manifold.

\section*{Acknowledgments}
We would like to thank Raunak Kumar, Rohan Jha, and our three reviewers for their thoughtful and thorough comments on improving our paper. In addition, we would like to thank our deep learning professor Joseph Redmon for inspiring this project. 

\bibliography{anthology,custom}
\bibliographystyle{acl_natbib}

\end{document}